\documentclass[runningheads]{llncs}
\usepackage{xcolor}
\usepackage{amsmath}
\usepackage[T1]{fontenc}
\usepackage{multirow}
\usepackage{graphicx}
\begin{document}
\title{Model Selection with a Shapelet-based Distance Measure for Multi-source Transfer Learning in Time Series Classification\thanks{This work was partially supported by MEXT-Japan (Grant No. 23K16949).}}

% \title{Contribution Title\thanks{Supported by organization x.}}
%
\titlerunning{Improving Time Series Transfer Learning}
\author{Jiseok Lee\and
Brian Kenji Iwana\orcidID{0000-0002-5146-6818}}
\authorrunning{Jiseok Lee and Brian Kenji Iwana}
% \institute{Department of Electrical Engineering and Computer Science, \\
% School of Engineering, Kyushu University, Fukuoka, Japan\\
% \email{jiseok.lee@human.ait.kyushu-u.ac.jp}\\
% \and 
\institute{Graduate School of Information Science and Electrical Engineering,\\
Kyushu University, Fukuoka, Japan \\
\email{jiseok.lee@human.ait.kyushu-u.ac.jp}\\
\email{iwana@ait.kyushu-u.ac.jp}}
\maketitle              % typeset the header of the contribution
\vspace{-0.6cm}
\begin{abstract}
Transfer learning is a common practice that alleviates the need for extensive data to train neural networks. It is performed by pre-training a model using a source dataset and fine-tuning it for a target task. However, not every source dataset is appropriate for each target dataset, especially for time series. 
In this paper, we propose a novel method of selecting and using multiple datasets for transfer learning for time series classification. 
Specifically, our method combines multiple datasets as one source dataset for pre-training neural networks. Furthermore, for selecting multiple sources, our method measures the transferability of datasets based on shapelet discovery for effective source selection.
While traditional transferability measures require considerable time for pre-training all the possible sources for source selection of each possible architecture, our method can be repeatedly used for every possible architecture with a single simple computation.
Using the proposed method, we demonstrate that it is possible to increase the performance of temporal convolutional neural networks (CNN) on time series datasets. 

\end{abstract}
\section{Introduction}
\label{sec:intro}
%time series
Neural networks have widespread usage in time series recognition. 
For example, temporal Convolutional Neural Networks (CNN)~\cite{Lecun_1998} have been shown to be effective across many time series domains~\cite{Wang_2017,bai2018empirical}. 
However, often, neural networks require large amounts of data~\cite{Ismail_Fawaz_2018,iwana2021an}.
Also, acquiring large amounts of annotated data can take time and effort.

%transfer learning
Several ideas exist to solve the problem of the requirement of large amounts of annotated data, such as transfer learning, self-supervised learning, data augmentation, etc.
In particular, transfer learning has become a popular method of initializing neural networks.
In transfer learning, to alleviate the need for data, neural networks can be trained on larger \textit{source} datasets and fine-tuned for \textit{target} datasets. 
In this way, the weights of the neural network can be trained to extract generalized features~\cite{Zhuang_2021} and be used for the target task. 
In the image recognition domain, transfer learning is a standard practice. 
For example, using pre-trained models trained with ImageNet~\cite{deng2009imagenet} in image recognition is standard practice.
However, for time series, transfer learning is still a budding field~\cite{Ismail_Fawaz_2018}.

% contribution
In order to realize an effective method of transfer learning for time series, we propose a combination of multi-source transfer learning with a novel shapelet-based distance measure used for dataset selection. 
Specifically, to increase the effectiveness of transfer learning and the source dataset's size, we propose a method of using multiple datasets for pre-training. 

However, selecting the datasets for transfer learning is complex.
Fawaz et al.~\cite{Ismail_Fawaz_2018} demonstrated that the choice of dataset for transfer learning for time series has a large effect on the effectiveness of transfer learning. 
Notably, only some datasets increased the accuracy of the model. 
Oftentimes, the accuracy was decreased when using an inappropriate dataset.

Thus, we propose a dataset distance-based measure to select the appropriate datasets for our multi-source pre-training. 
To do this, first, we extract discriminative shapelets using shapelet discovery~\cite{Ye_2009}. 
Next, a dataset distance measure is created by comparing the discriminative shapelets between the source and target datasets.
The idea is that datasets with similar discriminative shapelets would have similar features, thus leading to more effective transfer learning.

The contribution of this paper is as follows:
\begin{itemize}
    \item We propose a new method of predicting and selecting source datasets for transfer learning in temporal neural networks. This method uses extracted shapelet similarity between the target and possible source datasets.
    \item We create a transfer learning method that combines multiple sources into one super dataset. 
    \item We evaluate the proposed method on all 128 time series datasets from the 2018 UCR Time Series Archive (UCR Archive)~\cite{UCRArchive2018}.
    \item We provide code for easy transfer learning at 
    
    https://github.com/uchidalab/time-series-transferability
\end{itemize}

\section{Related Works}
\label{chap:bg}

\subsection{Transfer Learning for Time Series}
Transfer learning has been used for various applications in image recognition~\cite{Zhuang_2021}.
Furthermore, transfer learning has become the standard practice for training networks, as pre-trained weights are available for the most popular image recognition network architectures.

Conversely, transfer learning is less common for time series recognition and temporal neural networks~\cite{Ismail_Fawaz_2018} outside of Natural Language Processing (NLP). 
However, there have been a few works that demonstrate the usefulness of transfer learning in the time series domain~\cite{Ismail_Fawaz_2018,Weber_2021,Thompson_2022}.
Other examples include using transfer learning with health data~\cite{Clark_2022}, human activity recognition~\cite{an2023transfer}, prediction of internet load~\cite{Dridi_2021}, and fall detection~\cite{maray2023transfer}. 
De Souza et al.~\cite{desouza2021} proposed decomposing time series into shapelets and noise and using the decompositions to pre-train models. 
In comparison to our method, we use shapelets as a distance measure between datasets and not directly to train models.

For the source selection, several works demonstrated that source selection with dataset similarity, computed by using Dynamic Time Warping (DTW)~\cite{SAKOE_1990} distance, can be effective~\cite{Ismail_Fawaz_2018,ye2021implementing,lu2023multi}.

\subsection{Multi-Source Transfer Learning}
There have been a few works that use multiple source datasets for transfer learning.
For example, Yao and Doretto~\cite{Yao_2010} extend TrAdaBoost~\cite{Dai_2007}, a method of boosting transfer learning, to use with multiple sources. 
Huang et al.~\cite{Huang_2012} improve on this and propose SharedBoost for multiple source transfer learning. 
Multi-transfer~\cite{Tan_2013} combines multi-view and multi-source transfer learning. 
For multi-source transfer learning, Song et al.~\cite{Song_2017} use the conditional probability difference to weight source domains.

Multi-source transfer learning has also been used for time series data.
One of the typical methods of multi-source transfer learning is to use a preliminary classifier to select the sources. 
For example, for electroencephalogram (EEG) data, Jinpeng Li et al.~\cite{Li_2019} trained each source individually, and then they tested the target domain on each and selected the top-performing models. 
Ren et al.~\cite{Ren_2022} classify EEG data using a multi-source model. Huiming Lu et al.~\cite{lu2023multi} use an ensemble model to implement multi-source transfer learning for building energy prediction.

Also, Yao et al.~\cite{yao2024multi} use multi-source transfer learning with Variational Mode Decomposition to improve PM2.5 concentration forecasting while selecting sources using Euclidean Distance and Maximum Mean Discrepancy. 
Lotte and Guan~\cite{Lotte_2010} use a search algorithm to combine different datasets.
Senanayaka et al.~\cite{senanayaka2022similarity} used a similarity-based approach for multi-source transfer learning to generate a mixed domain of multiple sources and targets in the pre-training stage.

\section{Transferability Measure}

\subsection{Problem Setting}

\sloppy Given source dataset $\mathcal{S}=\{(\mathbf{s}_1,z_1), \dots, (\mathbf{s}_m, z_m), \dots, (\mathbf{s}_M, z_M)\}$, where $(\mathbf{s}_m, z_m)$ is the $m$-th pair of pattern $\mathbf{s}_m$ and respective label $z_m$, transfer learning aims to train a network $f$ with $\mathcal{S}$, so that it will be a practical starting point for target dataset $\mathcal{T}=\{(\mathbf{x}_1,y_1), \dots, (\mathbf{x}_n, y_n), \dots, (\mathbf{x}_N, y_N)\}$, where $(\mathbf{x}_n, z_n)$ is the $n$-th pair of pattern $\mathbf{x}_n$ and respective label $y_n$.
Unlike domain adaptation, there is no assumption that the task of the source and target datasets are related.

As Fawaz et al.~\cite{Ismail_Fawaz_2018} found, not all source datasets are useful for transfer learning with time series.
Thus, a suitable source dataset $\mathcal{S}$ for each target dataset $\mathcal{T}$ should be determined.
Under the problem setting, this determination should be performed before training $\mathcal{T}$, i.e., without exhaustively fine-tuning $\mathcal{T}$ on all possible datasets.

Two types of measures have been proposed to solve the source selection problem. 
One class of measures is to estimate the transferability of the pre-trained network $f$, and the other is to measure the similarity of the datasets. 

\subsection{Transferability Estimation}

Transferability estimation measures attempt to predict how effective transfer learning will be for model $f$, pre-trained on source dataset $\mathcal{S}$, for target dataset $\mathcal{T}$.
These methods first use models $f$ pre-trained on datasets $\mathcal{S}$ to predict target dataset $\mathcal{T}$. 
Specifically, prediction $f({x_n})$ is done using the data $x_n$ of $\mathcal{T}$ with the source labels $c\in C$ and the features $\mathcal{F}(x_n)\in\mathcal{F}$ are extracted from model $f$ trained by $\mathcal{S}$. 
While the source labels $C$ might be unrelated to the target task, the outputs are used for the transferability estimation. 
In other words, models trained for $\mathcal{S}$ are used as-is with dataset $\mathcal{T}$, and the amount of information inferred by the model is measured and used as a transferability estimation measure.
Some transferability estimation measures include, Log Expected Empirical Prediction (LEEP)~\cite{pmlr-v119-nguyen20b}, Negative Conditional Entropy (NCE)~\cite{Tran_2019}, Log Maximum Evidence (LogME)~\cite{pmlr-v139-you21b}, Transrate~\cite{pmlr-v162-huang22d}, and H-score~\cite{Bao_2019}.

For example, LEEP~\cite{pmlr-v119-nguyen20b} first predicts the target dataset $\mathcal{T}$ using trained $f$. 
LEEP is then calculated by:
\begin{equation}
    \mathrm{LEEP}(\mathcal{S}, \mathcal{T}) = \frac{1}{N}\sum^{N}_{n=1}\mathrm{log}\left(\sum_{c\in C}\hat{P}(y_n|c)f(x_n)\right),
\end{equation}
where $\hat{P}(y_n|c)$ is the empirical conditional distribution calculated by:
\begin{equation}
    \hat{P}(y_n|c)=\frac{\hat{P}(y_n,c)}{\hat{P}(c)}=\frac{\frac{1}{N}\sum_{n:y_n=c}f(x_n)}{\frac{1}{N}\sum^N_{n=1}f(x_n)}.
\end{equation}
LEEP is the average log-likelihood of the prediction of $\mathcal{T}$ in trained network $f$ multiplied by $\hat{P}(y_n|c)$ for each source class $c$.

\subsubsection{Dataset Similarity Measure for Source Selection}

As an alternative to transferability estimation, dataset similarity can also be used to predict transferability. 
The previous methods are helpful because only the pre-trained network $f$ and not the original dataset $\mathcal{S}$ is needed to calculate transferability.
However, unlike image recognition, standard models with downloadable weights for time series recognition are lacking.
Therefore, requiring pre-trained networks is a detriment because it requires training many networks before pre-training the actual network for the task.

Conversely, measuring the distance between datasets only requires access to the datasets. 
Following this, Fawaz et al.~\cite{Ismail_Fawaz_2018} proposed to use DTW~\cite{SAKOE_1990} between representative time series patterns from each class in the target and source datasets. 
The representative time series is the average time series of each class found by DTW Barycenter Averaging (DBA)~\cite{Petitjean_2011}. 
They defined the distance between datasets as the distance between the most similar average time series from each dataset. 
In this paper, we define this method as DBA-DTW.
By using the dataset-based distance measure, the appropriate source dataset can be selected for the target dataset. 
The benefit of this and the proposed method is that no initial trained model is required to measure transferability.

\section{Multi-Source Transfer Learning}
\begin{figure}[tb]
\centering
\includegraphics[width=0.37\columnwidth,trim={0 5px 0 0},clip]{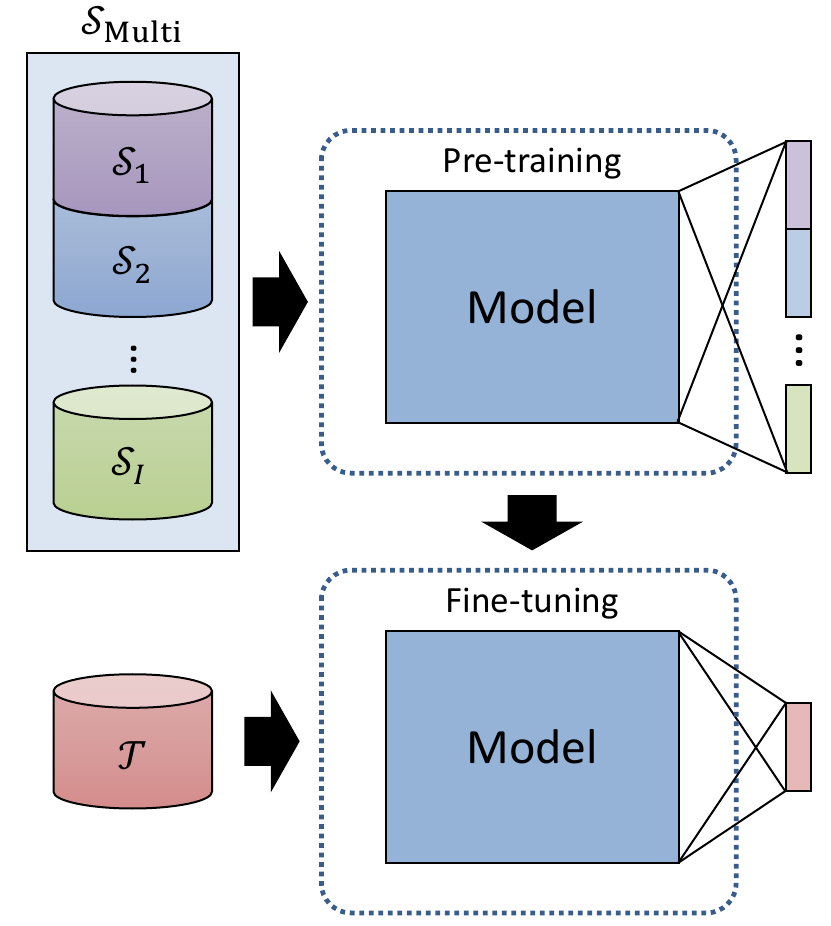}
\caption{An illustration of our multi-source transfer learning. Source datasets $\mathcal{S}_i$ are selected using a transferability measure, and the neural network is pre-trained. The trained weights are then fine-tuned using target dataset $\mathcal{T}$.}
\label{fig:model}
% \vspace{-0.2cm}
%
\end{figure}

We propose a simple yet effective method of combining multiple source datasets for transfer learning. 
As shown in Fig.~\ref{fig:model}, to perform multi-source pre-training, we compile multiple source datasets $\mathcal{S}_1,\dots,\mathcal{S}_i,\dots,\mathcal{S}_I$ into one super dataset $\mathcal{S}_{\mathrm{Multi}} = \{\mathcal{S}_1,\dots,\mathcal{S}_i,\dots,\mathcal{S}_I\}$. 
In order to transfer knowledge from a model trained with multiple sources, the datasets are pre-processed so that they have the same number of time steps as the target dataset. 
Note that by resampling the source datasets, the features and characteristics of the datasets might not be preserved. 
However, this fact is optional because the purpose of the pre-training is to acquire a robust set of initial weights for transfer learning and not the classification accuracy of the source datasets.
In addition to resampling, $\mathcal{S}_{\mathrm{Multi}}$ is balanced so that each sub-dataset has the same number of time series. 
To balance $\mathcal{S}_{\mathrm{Multi}}$, oversampling is performed while preserving the class ratios. 
This is done due to the discrepancy in the size of possible source datasets; it ensures that every dataset has an equal contribution to the transfer learning.

As shown in Fig.~\ref{fig:model}, in order to train a network with multiple datasets, the one-hot vector of the ground truth of each source dataset $\mathcal{S}_i$ are concatenated, and the output nodes are extended accordingly. 
In this way, the network is then trained using $\mathcal{S}_{\mathrm{Multi}}$ for a classification task using all of the classes from all of the source datasets.
The result is the ability to pre-train a network with a larger dataset with a larger number of classes.

After training the network, fine-tuning can be performed as typical transfer learning does.
% After training the network, typical transfer learning can be performed. 
The weights of the trained network can be used as an initialization for a target dataset and fine-tuned for a specific task.
While the experimental results use temporal CNNs, there is no theoretical limitation on the type of neural network used.used. %Remove for version2

\section{Shapelet Similarity-based Source Selection}

In order to use the proposed transfer learning method effectively, some source datasets need to be selected. 
However, as mentioned, selecting the source datasets needs to be performed. 
Thus, we propose a new method of measuring the transferability of networks through a novel dataset similarity measure using discriminative shapelets.

\subsection{Shapelet}
\begin{figure}[tb]
\centering
\includegraphics[width=0.45\columnwidth]{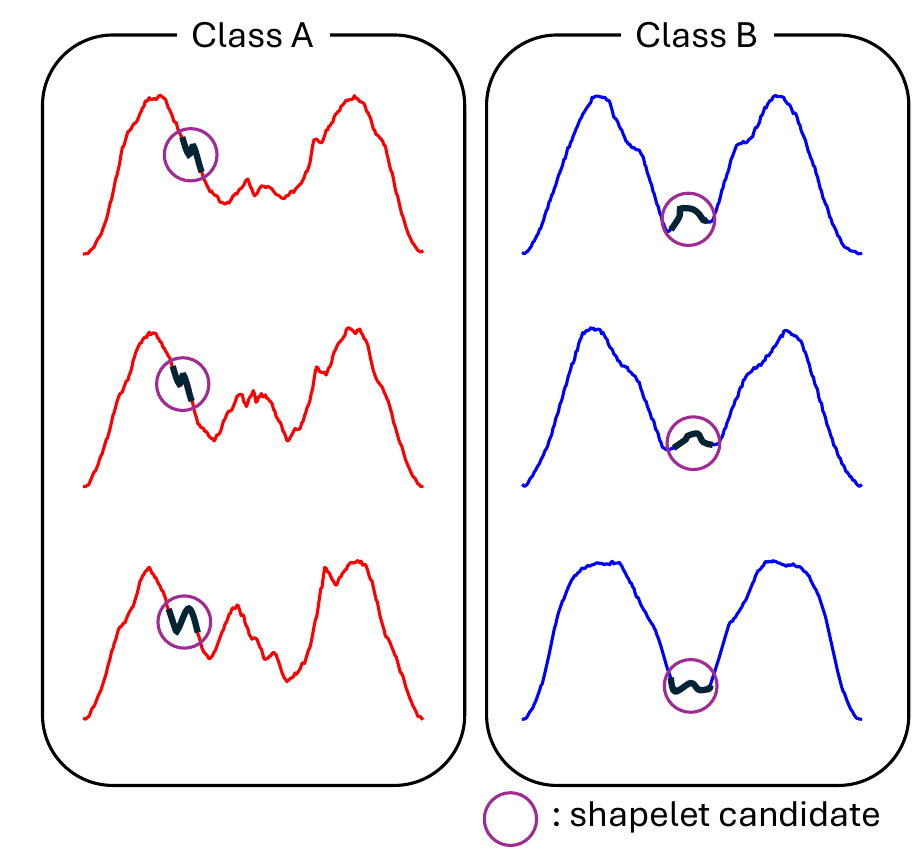}
% \vspace{-0.4cm}
\caption{Examples of shapelets from the Arrowhead dataset. The left and right figures are three time series patterns from the same classes.}
\label{fig:shapelet}
% \vspace{-0.2cm}
\end{figure}

A shapelet refers to a subsequence extracted from time series data that are maximally representative of a class~\cite{Ye_2009}. 
These subsequences are intended to encapsulate fundamental patterns or discriminative features within a class.
For example, the circled subsequences in Fig.~\ref{fig:shapelet} are well discovered within a class and represent differences between the two classes. 
In the figure, the circled shapelets are segments of the time series unique to each class.

\subsection{Matrix Profile for Shapelet Discovery}

Since a shapelet can be any subsequence from a time series, finding the maximum representative shapelet would be too costly with brute force. 
In order to overcome this issue, Matrix Profile~\cite{Yeh_2016} has been shown to find discriminative shapelet candidates efficiently. 

Matrix Profile is an algorithm that represents a time series based on the distances between subsequences of that time series and their nearest neighbor. 
Specifically, given an ordered list of all subsequences $\mathcal{A}$ of a single time series $\mathbf{t}$, Matrix Profile $\mathbf{p}$ is a sequence that holds the distances between each subsequence $\mathcal{A}_r$ to its nearest neighbor, or:
\begin{equation}
\label{eq:differences}
    \mathbf{p} = ||\mathcal{A}_{1} - \mathcal{E}_{1}||, \dots, ||\mathcal{A}_{r} - \mathcal{E}_{r}||, \dots, ||\mathcal{A}_{R} - \mathcal{E}_{R}||,
\end{equation}
where $\mathcal{E}_r$ is the nearest subsequence of $\mathcal{A}$ to the respective $\mathcal{A}_r$ and $||\cdot||$ is the sum of the pair-wise Euclidean distances between each element in the subsequences.
By finding Matrix Profile $\mathbf{p}$ using time series $\mathbf{t}$, Matrix Profile can be used as a fast motif and discord discovery method.

In order to use Matrix Profile for shapelet discovery, a few modifications are performed.
First, given a dataset $\mathcal{S}$, all of the time series of each class $c$ are concatenated into a single time series $\mathbf{t}^{(c)}$. 
For example, given class 1 and class 2, a time series $\mathbf{t}^{(1)}$ and $\mathbf{t}^{(2)}$ are created.
Then, instead of just calculating Matrix Profile using the nearest neighbors of $\mathcal{A}^{(1)}$ with itself in \eqref{eq:differences}, a Matrix Profile calculation is made for each combination of the two classes, or $\mathbf{p}^{(1,1)}$, $\mathbf{p}^{(1,2)}$, $\mathbf{p}^{(2,2)}$, and $\mathbf{p}^{(2,1)}$. 
Finding the highest values in the differences $\mathbf{p}^{(1,2)}-\mathbf{p}^{(1,1)}$ and $\mathbf{p}^{(2,1)}-\mathbf{p}^{(2,2)}$ will identify the maximally representative shapelet candidates for class 1 and 2, respectively. 
Because this is only compatible with two-class classification, we extend shapelet discovery via Matrix Profile using a one-versus-all approach for each class.

\subsection{Shapelet Similarity-based Source Selection}
\begin{figure}[tb]
\centering
\includegraphics[width=0.75\columnwidth]{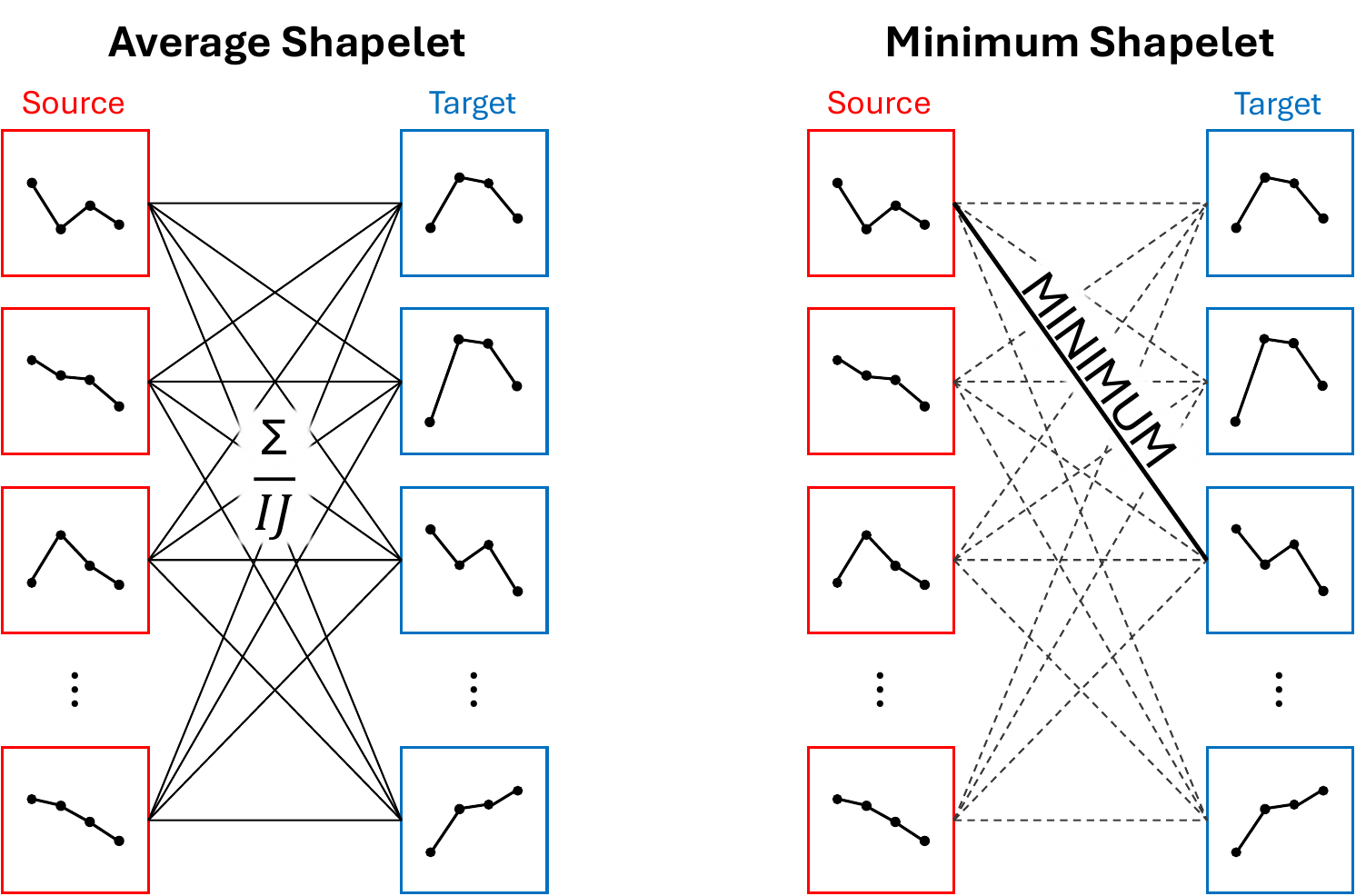}

\caption{Overview of Average Shapelet and Minimum Shapelet.}
\label{fig:shapelet_method}
% \vspace{-0.2cm}
\end{figure}

Now that the representative shapelets $\mathcal{P}^{(\mathcal{S})}$ and $\mathcal{P}^{(\mathcal{T})}$ can be found for each source dataset $\mathcal{S}$ and target dataset $\mathcal{T}$, respectively, we use them as a basis for a dataset distance measure.
% We propose three shapelet distance measure schemes. 
We propose two shapelet distance measure schemes, Average Shapelet, and Minimum Shapelet distances. 
Fig.~\ref{fig:shapelet_method} represents an overview of Average Shapelet and Minimum Shapelet.

\textbf{Average Shapelet} takes the average distance for each combination of $\mathcal{P}^{(\mathcal{S})}_i$ and $\mathcal{P}^{(\mathcal{T})}_j$, or:
\begin{equation}
D_\mathrm{as}=\frac{1}{IJ} \sum_{i,j} ||\mathcal{P}^{(\mathcal{S})}_i - \mathcal{P}^{(\mathcal{T})}_j||,
\end{equation}
where $\mathcal{P}^{(\mathcal{S})}_i$ and $\mathcal{P}^{(\mathcal{S})}_j$ are the $i$-th and $j$-th shapelet in $\mathcal{P}^{(\mathcal{S})}$ and $\mathcal{P}^{(\mathcal{T})}$, respectively. 
By using the average distance between shapelets, this measure compares all of the discriminative features of the datasets simultaneously.
The general idea of this measure is that if all of the shapelets of the datasets are similar, then the datasets might be similar.

\textbf{Minimum Shapelet} is defined as the distance between the most similar pair of shapelets of $\mathcal{S}$ and $\mathcal{T}$, or:
\begin{equation}
D_\mathrm{ms}=\min_{i,j} ||\mathcal{P}^{(\mathcal{S})}_i - \mathcal{P}^{(\mathcal{T})}_j||.
\end{equation}
Instead of measuring all of the features, this measure allows the distance calculation to ignore features that might be specific to a dataset.

\section{Experimental Result}
\subsection{Dataset}
The experiments were conducted using all of the UCR Archive~\cite{UCRArchive2018}, which consists of 128 datasets.
We use the predetermined training and test set split provided by the archive.
Also, no pre-processing was performed except for resizing datasets through Gaussian smoothing with different lengths.

\subsection{Settings and Architecture}
%CNN
For the experiment, we adopted a 1-dimensional CNN model based on the VGG architecture~\cite{simonyan2015deep}.
% and a 1-dimensional transformer model~\cite{vaswani2017attention}. 
The convolutional network used three blocks of convolutional layers and a pooling layer. The first block has two convolutional layers of size 3, and the subsequent blocks have three convolutional layers. 
Max pooling is used with the first two blocks, and global average pooling (GAP) is used with the final block. 
While GAP is not required for the proposed method, it is required for LEAP, NCE, H-score, Transrate, and LogMe due to having different-sized datasets; thus, we used it for all evaluations.

For training, we pre-train the network for 10,000 iterations with Adaptive Moment Estimation (Adam) optimizer~\cite{kingma2014adam} with an initial learning rate of 0.0001.
For transfer learning, we then fine-tune the network for 5,000 iterations. 
The batch size is set to 32 for both pre-training and fine-tuning.
For statistical validity, we fine-tuned the model three times in order to have the mean of the three models' performances.
Since the traditional transferability measures, LEEP, NCE, LogME, Transrate, and H-Score, require a trained network, an initial network to calculate the transferability is trained for 5,000 iterations.

There are two hyperparameters associated with the shapelet discovery by Matrix Profile. 
First, we use a fixed shapelet size of 15 because this is the largest possible size on the UCR Archive's smallest dataset. 
Next, we use the top 10 shapelet candidates per class.
% , 5, and 3

\subsection{Comparative Evaluation}
To evaluate the proposed method, we compared it to not using transfer learning, to using transfer learning using a dataset selected by a shapelelt-based distance measure, and to using a dataset selected by the other transferability metrics.
The comparative measures used for source selection include using DTW between DBA class representatives (DBA-DTW)~\cite{Ismail_Fawaz_2018}, LEEP~\cite{pmlr-v119-nguyen20b}, NCE~\cite{Tran_2019}, Transrate~\cite{pmlr-v162-huang22d}, LogME~\cite{pmlr-v139-you21b}, and H-score~\cite{Bao_2019}.

% without transformer
\begin{table}
    \caption{Average test accuracy.}
    % \vspace{-4mm}
    \label{tab:results}
    \centering
    \begin{tabular}{lcc}
    \hline
    Method & Accuracy (\%) \\
    \hline
    No Transfer Learning (TL) & 74.18\\
    \hline
    TL w/ Random Source& 76.89\\
    TL w/ DBA-DTW & 78.85\\
    TL w/ H-Score & 77.24\\
    TL w/ LEEP & 76.43\\
    TL w/ LogME & 79.59\\
    TL w/ NCE & 78.92\\
    TL w/ Transrate & 77.22\\
    \hline
    Proposed w/ Average Shapelet (10 shapelets) & 79.91\\
    Proposed w/ Minimum Shapelet (10 shapelets)  & \textbf{80.25}\\ 
    \hline
    \end{tabular}
\end{table}

The results of the experiments are shown in Table~\ref{tab:results}. 
% In the table, the comparison methods, and TL w/ AS and MS use the proposed shapelet-based measure with multi-source transfer learning using $I=14$ and $I=16$, where $n$ cand. indicates the number of candidates for shapelet discovery.
In the table, TL represents typical transfer learning that selects a single source dataset, and MTL represents our proposed method, Multi-source Transfer Learning.
Compared to the model with random initialized weights, Multi-source Transfer Learning with a Minimum Shapelet of 10 candidates showed the highest improvement.
On the other hand, the Average Shapelet showed lower improvement than the Minimum Shapelet; however, it is still better than other comparative methods.

\begin{figure}[tb]
\centering
\includegraphics[width=0.65\columnwidth,clip]{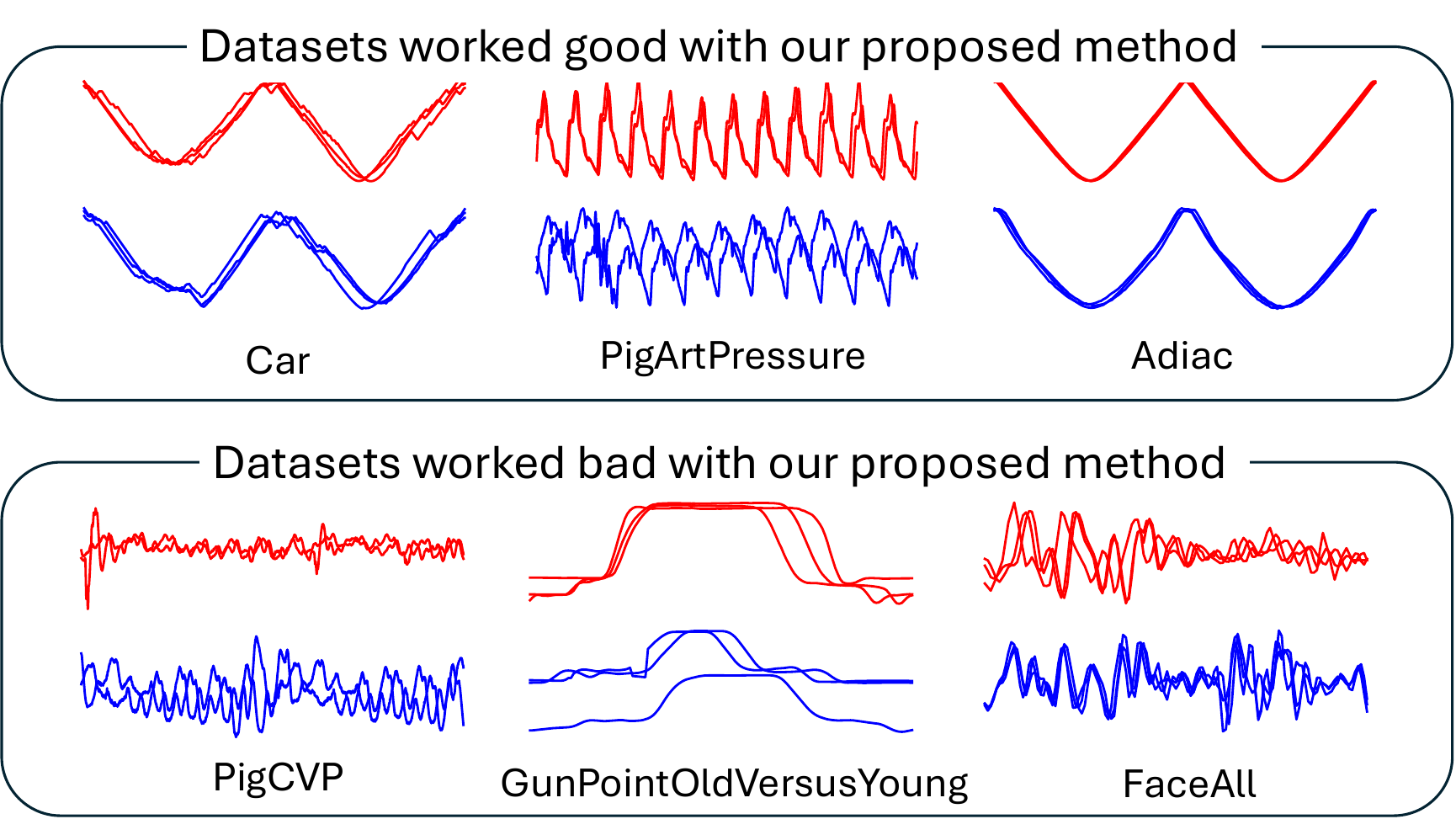}
\caption{Sample plot of datasets of UCR Archive. Random three samples are plotted from two classes. The upper three datasets performed better with our proposed method, and the lower three datasets showed adverse effects by adopting our proposed method.}
\label{fig:sam}
% \vspace{-0.2cm}
\end{figure}

Additionally, we looked into the datasets that showed better and worse performance by adopting our proposed method.
Fig.~\ref{fig:sam} shows sample plots of the datasets.
From the sample plot, we can affirm that our proposed method works better with datasets with more clear features.
Also, the datasets with more noise showed worse performance than our proposed method.
This trend is due to our proposed method focusing on shapelet similarity.

\begin{figure}[tb]
\centering
\includegraphics[width=1\columnwidth,clip]{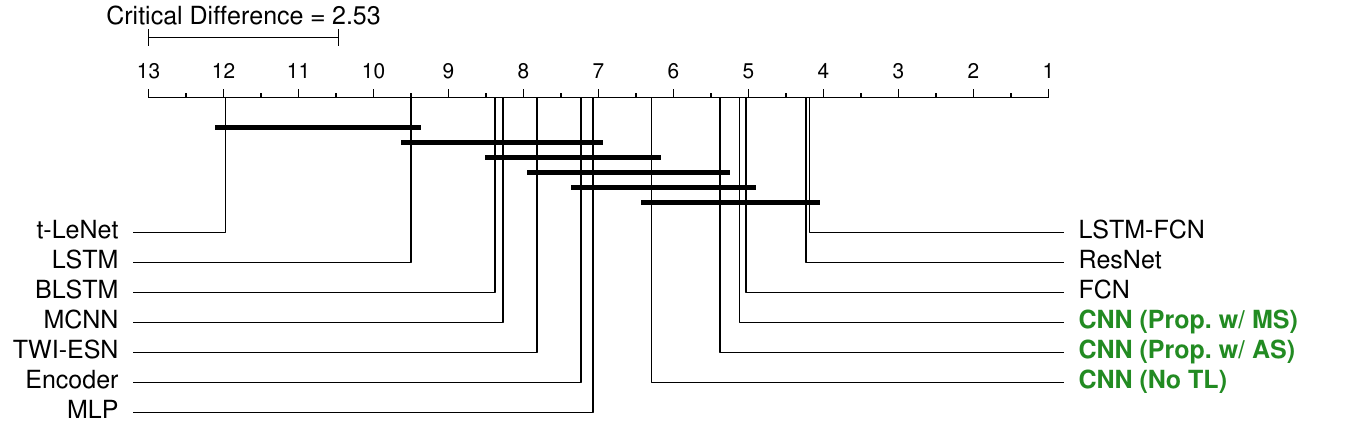}
\vspace{-0.6cm}
\caption{A Nemenyi post-hoc test diagram. The proposed methods are in green. The numbers indicate the average rank when tested on all 128 datasets.}
\label{fig:nem}
% \vspace{-0.2cm}
\end{figure}

To compare the proposed method to other methods found in the literature, we compare it against reported results on the UCR Archive from methods that use neural networks. 
Fig.~\ref{fig:nem} is a Nemenyi post-hoc test diagram comparing the proposed methods, the base models, and various reported evaluations. 
The comparison models include a temporal Residual Network~(ResNet)~\cite{Wang_2017}, Fully Convolutional Network (FCN)~\cite{Wang_2017}, MLP~\cite{Wang_2017}, Multi-scale CNN~(MCNN)~\cite{cui2016multi}, Time Warping Invariant Echo State Network~(TWI-ESN)~\cite{tanisaro2016time}, Time Le-Net~(t-LeNet)~\cite{le2016data}, universal Encoder~\cite{serra2018towards}, LSTM~\cite{hochreiter1997long}, BLSTM~\cite{Schuster_1997}, LSTM-FCN~\cite{karim2017lstm}. The models were evaluated by Wang et al.~\cite{Wang_2017}, Ismail Fawaz et al.~\cite{ismail2019deep}, and Iwana and Uchida~\cite{iwana2021an}.
The figure shows that the networks used by the proposed method are comparable to the other state-of-the-art neural networks on the same datasets.

\section{Discussion}

\subsection{Ablation Study}
\begin{figure}[tb]
\centering
\includegraphics[width=\columnwidth,clip]{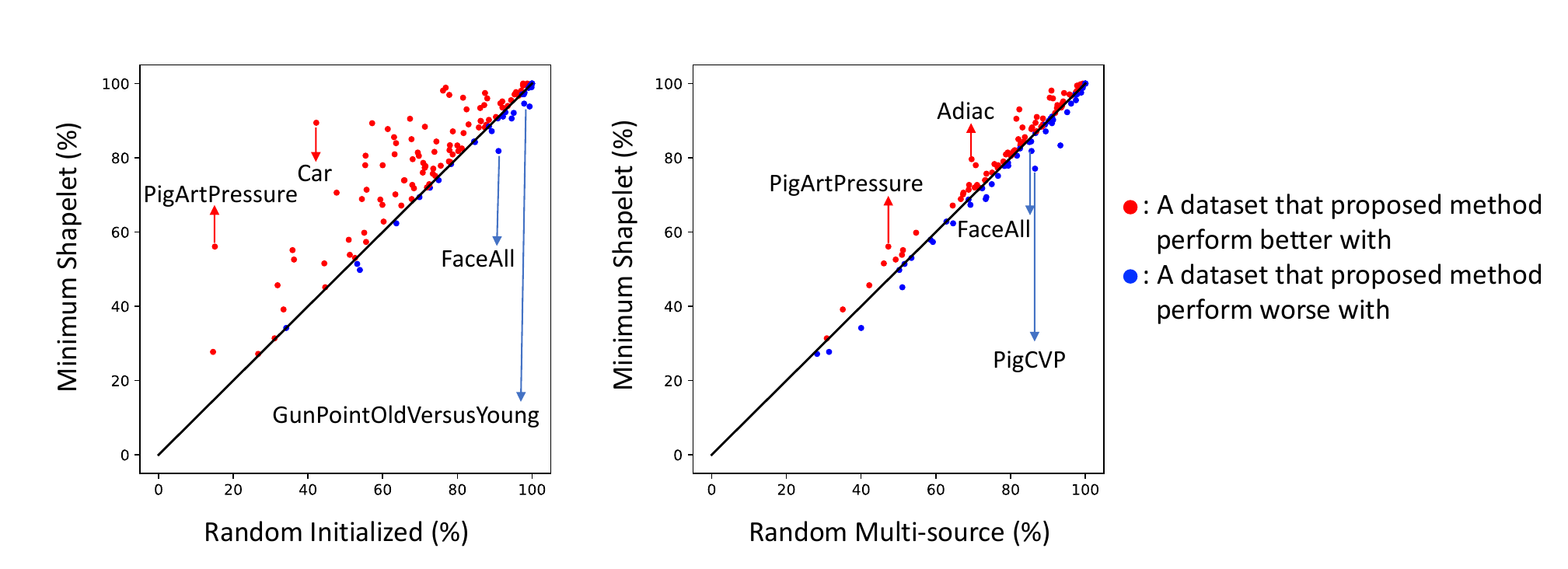}
\vspace{-0.5cm}
\caption{Ablation study of our proposed method. 
The Y-axis represents the classification accuracy of our proposed method with 14 sources selected by Minimum Shapalet. The X-axis of the left and right figures represent the classification accuracy of random initialized models and proposed multi-source transfer learning with 14 random sources.
}
\label{fig:ab}
% \vspace{-0.2cm}
\end{figure}

We compared the classification performances with multi-source pre-training with shapelet-based source selection, multi-source pre-training with random source selection, and without transfer learning.
Fig.~\ref{fig:ab} compares the performances of each dataset of the UCR Archive.
Our proposed method showed significantly better results than the random initialized model and multi-source transfer learning with random source selection.
Specifically, the result on the right figure demonstrates that our proposed shapelet similarity-based source selection is effective for source selection on multi-source transfer learning.
Also, according to the t-test, our proposed method is effective with most datasets, p < 0.001.

\subsection{Number of Datasets}
In order to examine how the hyperparameters for the proposed method affect performance, we examined the number of shapelet candidates, the number of source datasets, and the two shapelet similarity-based distance measures.
For the number of shapelet candidates, we discovered three, five, and ten shapelet candidates for every dataset. For the number of source datasets, we selected 1, 2, 4, 6, 8, 10, 12, 14, 16, 18, and 20 datasets according to shapelet-based distance measures. Finally, we examined the Average Shapelet and the Minimum Shapelet for the shapelet-based distance measure.

\begin{figure}[tb]
\centering
\includegraphics[width=0.95\columnwidth,clip]{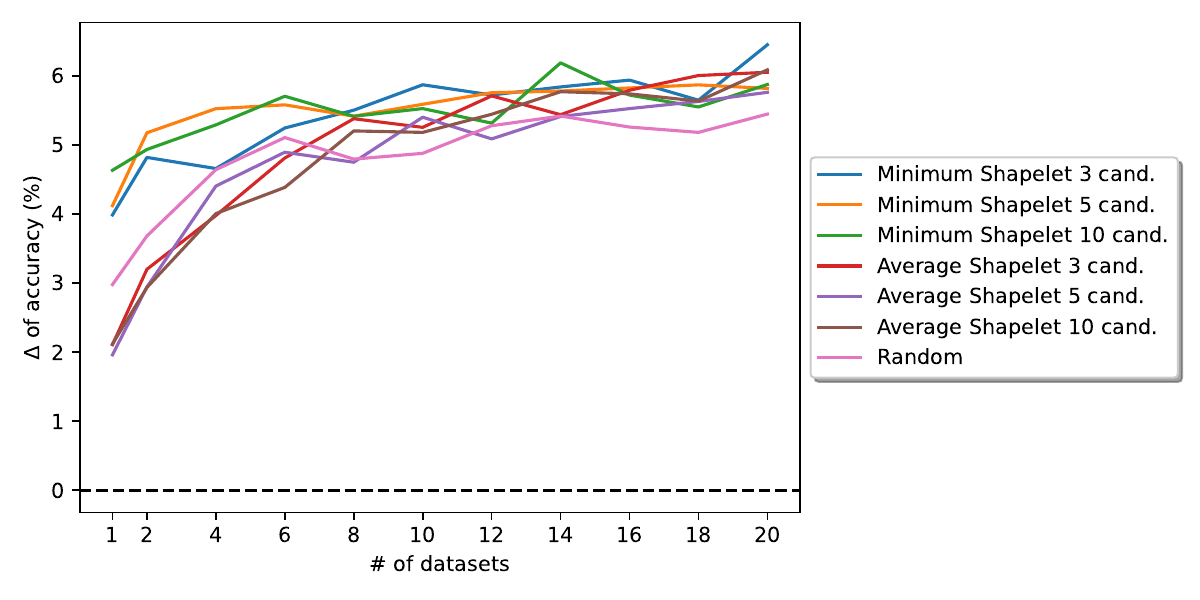}
\vspace{-0.5cm}
\caption{Experimental result of examining hyperparameters.}
\label{fig:exp1}
% \vspace{-0.2cm}
\end{figure}

The experimental results in Fig.~\ref{fig:exp1} showed better performance with more sources for all selection methods, including Random multi-source selection.
Thus, increasing the number of datasets using our multi-source transfer learning method effectively increases the effect of pre-training while alleviating the risk of negative transfer.
% selected according to DBA-DTW and shapelet-based distance measurements than fewer source datasets. 
However, there was no significant difference when the number of datasets was more than 10. 
This implies a diminishing returns effect with the number of datasets. 
Thus, there is a limit to how many datasets should be combined. 
However, one possible reason for the diminishing returns is due to our training scheme. 
We set the number of pre-training iterations to a fixed number, no matter how many datasets and data samples were added for a fair comparison.
Therefore, it might be possible to increase the accuracy further using multiple datasets with more iterations.

\subsection{Similarity Based Source Selection}

Generally, with a small-scale target dataset, pre-training with a closely related source dataset allows for more efficient training while reducing the risk of overfitting. However, many time series tasks such as EGG, Speeches, and Gestures have different features and usually contain a small number of patterns. Thus, selecting a source dataset for time series transfer learning is often a big challenge, and transfer learning with non-related tasks is usually not helpful.

According to the experimental result, as depicted in Table~\ref{tab:results}, selecting a source dataset based on dataset ranking metrics of DBA-DTW and Minimum Shapelet resulted in superior performance compared to a random source selection and even to the classic transferability measures. Therefore, for time series classification, where lack of data is a frequent challenge, measuring time series similarity can serve as a valuable indicator of transferability.

\subsection{Computational Time}
Regarding computational time, shapelet similarity-based source selection has a huge benefit compared to other transferability estimation metrics. Shapelet similarity-based dataset ranking has two steps of calculation: one step is to generate shapelet, and the other is to calculate the similarity among generated shapelets. Thus, the shapelet similarity-based source selection requires $O(n^2 + w)$; $O(n^2)$ to generate shapelet and $O(w_s)$ to calculate the similarity among shapelets, where $n$ is the length of the dataset and $w_s$ is the length of shapelet. The DBA-DTW also has the benefits of computational time compared to the other transferability estimation measures; however, it is not like the shapelet similarity-based method. DBA-DTW requires $O(i\cdot n^2+w_d^2)$; $O(i\cdot n^2)$ to generate prototypes through DBA and $O(w_d^2)$ to calculate distance among generated shapelets, where $w_d$ is the length of a prototype of DBA.

The other transferability estimation measures, such as LEEP, require a pre-trained model to measure transferability. Training a model requires significant time, and they have a disadvantage in that they need to re-calculate the transferability when the model architecture is changed. 
In this research, it is required to train $128 \times 128$ models for evaluation. However, our proposed method requires no re-calculation even though the model architecture has been changed. 
Thus, unlike transferability estimation measures, additional datasets can be calculated quickly and used for other tasks. 

\section{Conclusion}
\label{chap:conc}
In this paper, we suggest using transfer learning for temporal neural networks using a proposed multi-source pre-training. 
Specifically, we demonstrate that by combining multiple datasets into a super dataset using pre-processing and adjusting the classification task with the concatenation of the classes, it is possible to pre-train a network using a large amount of data and classes. 

Furthermore, in order to select appropriate datasets out of the large number of possible datasets that exist, we propose a new transferability measure based on shapelets. 
Our novel method calculates the distance between datasets using a shapelet similarity. 
The shapelet-based distance compares the class-discriminative shapelets between classes of the target dataset and the source dataset.
We demonstrate that by using multi-source transfer learning with our shapelet similarity-based source selection, it is possible to increase the time series classification accuracy with little downside.

In future work, we will investigate the combinations of source datasets to optimize our proposed method further. 
We hope to contribute to the time series classification community by continuing to expand upon these techniques.

\bibliographystyle{splncs04}
\bibliography{mybibliography}

\end{document}